\documentclass[10pt, conference, compsocconf]{IEEEtran}

\usepackage{cite}
\usepackage[pdftex]{graphicx}
\graphicspath{ {Images/} }

\usepackage[cmex10]{amsmath}
\usepackage{amsfonts}
\usepackage{amssymb}
\usepackage[caption=false,font=footnotesize]{subfig}
\usepackage{url}
\usepackage{caption}
\usepackage{tabularx}
\usepackage{multirow}
\usepackage{xcolor}

\makeatletter
\renewcommand\abstract[1]{%
	\def\@abstract{%
		\centerline{{\large\bf Abstract}}
		\noindent
		#1}}

\newcommand{\pname}[1]{{{LCANet}}{#1}}

\newcommand{\figbox}[2][.8in]{%
	\fbox{%
		\vbox to .6in{%
			\vfil
			\hbox to #1{%
				\hfil
				#2%
				\hfil}%
			\vfil}}}

\begin{document}

\IEEEoverridecommandlockouts

\title{\LARGE \bf
		\pname{}:	End-to-End Lipreading with Cascaded Attention-CTC
}
	
	\author{\parbox{16cm}{\centering
    {\large Kai Xu$^1$, Dawei Li$^2$, Nick Cassimatis$^2$, Xiaolong Wang$^2$*
    \thanks{*Corresponding author.} 
    \thanks{Work was done during Kai's internship at Samsung Research America.}
    \thanks{978-1-5386-2335-0/18/\$31.00~\copyright{}2018 IEEE \hfill}
    }\\
    {\normalsize
    $^1$ Arizona State University, Tempe, AZ 85281, USA\\
    $^2$ Samsung Research America, Mountain View, CA 94043, USA
    }}
}
	
	
	\maketitle

	\begin{abstract}

Machine lipreading is a special type of automatic speech recognition (ASR) which transcribes human speech by visually interpreting the movement of related face regions including lips, face, and tongue. Recently, deep neural network based lipreading methods show great potential and have exceeded the accuracy of experienced human lipreaders in some benchmark datasets. However, lipreading is still far from being solved, and existing methods tend to have high error rates on the wild data. In this paper, we propose \pname{}, an end-to-end deep neural network based lipreading system. \pname{} encodes input video frames using a stacked 3D convolutional neural network (CNN), highway network and bidirectional GRU network. The encoder effectively captures both short-term and long-term spatio-temporal information. More importantly, \pname{} incorporates a cascaded attention-CTC decoder to generate output texts. By cascading CTC with attention, it partially eliminates the defect of the conditional independence assumption of CTC within the hidden neural layers, and this yields notably performance improvement as well as faster convergence. The experimental results show the proposed system achieves a 1.3$\%$ CER and 3.0$\%$ WER on the GRID corpus database, leading to a 12.3$\%$ improvement compared to the state-of-the-art methods.

	

\end{abstract}  
	
	
	\IEEEpeerreviewmaketitle

    \section{Introduction}
	
	Lipreading, the ability to understand what people are saying from only visual information, has long been believed to be a superpower as shown in some science fiction movies. The recent development of deep learning techniques has begun to unveil the secret behind it and demonstrated the potential to make it available for commons. Compared to the acoustic-based speech recognition technique, lipreading solves a much broader range of practical problems, such as aids for hearing-impaired persons, speech recognition in a noisy environment, silent movie analysis, etc. 
	

	Lipreading is a challenging problem because the differences among some of the phonemes or visemes are subtle (e.g., the visemes corresponding to ``p'' and ``b'') although their corresponding utterances can be easily distinguished. Such fine distinction is the major obstacle for humans to read from lips and as reported previously only around 20\% reading accuracy can be achieved \cite{hilder2009comparison}. On the other hand, recent work shows that machine lipreading has got remarkable progress especially with the advancement of deep learning techniques \cite{lipnet:Petridis, lipnet:Nando, lipreading:Wand, lipwild:Chung, lipcnn:Noda}. 

	\begin{figure}[t]
		\centering
		\includegraphics[height=3.3cm]{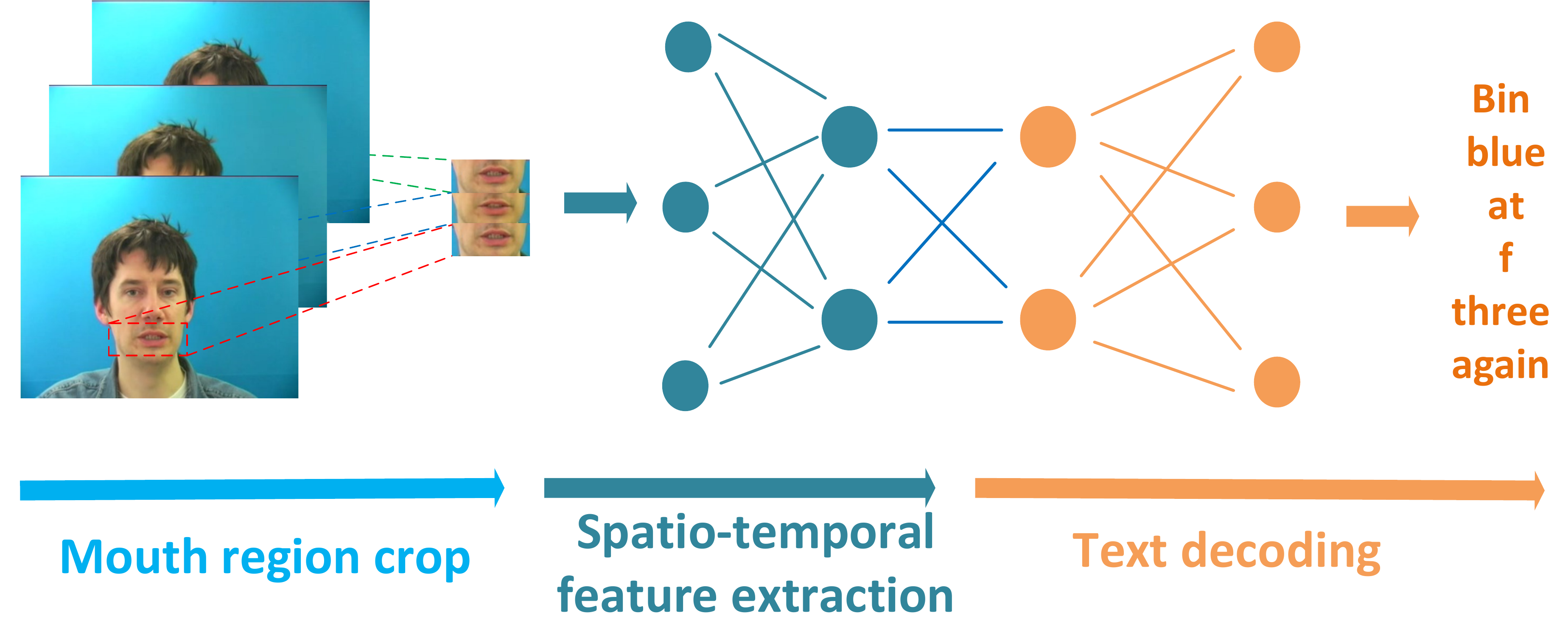}
		\vspace{0in}
		\caption{The proposed end-to-end lipreading system includes three major steps. (1) The mouth regions are cropped from the input video frames from aligned faces. (2) An encoder extracts spatio-temporal features from the input sequence. (3) The cascaded attention-CTC decoder generates text from encoded hidden features.}
		\label{fig:intro}
	\end{figure}
	
	Deep learning techniques have been the cornerstone for recent success \cite{geng2017novel, zhang2017learning,miao2016cnn,Tome_2017_CVPR,peng2017reconstruction} of many traditionally hard machine learning tasks \cite{wu2015fully,wu2016robust,ding2016articulated,meng_Face, wang2013automatic,7884622}. Generally, deep learning based lipreading methods follow the state-of-the-art sequential modeling solutions that have been widely applied to problems like acoustic based speech recognition \cite{deepspeech2,eesen:Miao, asr:Zhang} and neural machine translation \cite{attention,sennrich2015neural,gehring2017convolutional,vaswani2017attention}. Two most widely used approaches are Connectionist Temporal Classification (CTC) \cite{ctc:Graves} and the attention-based sequence to sequence (seq2seq) model \cite{attention}.
	
	CTC approach works well on acoustic-based speech recognition. For lipreading problem, \cite{lipnet:Nando} uses the CTC loss function \cite{ctc:Graves} to train an end-to-end deep neural network, named ``LipNet", and the model outperforms experienced human lipreaders on the GRID benchmark dataset \cite{GRID}. However, CTC loss function assumes conditional independence of separate labels (i.e., the individual character symbols), and each output unit is the probability of observing one particular label at a time. Therefore, although CTC is applied on top of Recurrent Neural Networks (RNNs), it tends to focus too much on local information (nearby frames) \cite{lipwild:Chung}. This may work well for acoustic phonemes, but is not well fit for predicting visemes which require longer context information to discriminate their subtle differences.


	
	 Attention-based seq2seq models were initially introduced for neural machine translation \cite{attention}, for which the input sequence and output sequence can be chronologically inconsistent. Such kind of model has also been adapted to solve the lipreading problem \cite{lipwild:Chung}: an encoder encodes the input frames (optionally with audios) into hidden features which are fed into an attention-based decoder to obtain the text output. By eliminating the conditional independence assumption, the attention-based method achieves better performance than CTC. However, recent work \cite{chen2017progressive,erdogan2016multi} shows that the attention model is too flexible to predict proper alignments for speech recognition, and is hard to train from scratch due to the misalignment on longer input sequences.
	
	In this paper, we propose a cascaded attention-CTC deep learning model, \pname{}, for end-to-end lipreading. \pname{} incorporates an encoder that is composed of 3D convolution, highway network and bi-directional GRU layers, which is able to effectively capture both short-term and long-term spatio-temporal information from the input video frames. To deal with the conditional independence assumption of the CTC loss, the encoded features are fed into a cascaded attention-CTC decoder. The proposed decoder explicitly absorbs information from the longer context from the hidden layers using the attention mechanism. Extensive evaluation result shows that \pname{} achieves the state-of-the-art performance and faster convergence than existing methods.

	The contribution of this paper is as follows:
	\begin{itemize}
		\item We propose an end-to-end lipreading deep neural network architecture, \pname{}, which relies purely on the visual information. 
		\item \pname{} leverages a cascaded attention-CTC decoder to generate text from encoded spatio-temporal features. Such design partially eliminate the defect of the conditional independence assumption in the vanilla CTC-based method.
		\item  We demonstrate the performance improvement compared with existing work. The benefits of the cascaded attention-CTC decoder are also discussed.
	\end{itemize}
	

    \section{Related Work}

\subsection{Deep Learning for Speech Recognition}

Recent advancement of speech recognition technique is primarily due to the prosperity of deep learning research \cite{dahl2012context, asr:Graves, hinton2012deep, deepspeech2, bahdanau2016end, las:William, eesen:Miao}, and the reported speech recognition error rate is already below that of a human. This makes speech a convenient input method for computers as well as mobile or embedded devices. Together with the advancement of natural language understanding techniques \cite{kim2016character, nlp2, nlp1,nlp3} and recent software \cite{deeprebirth,DBLP:journals/corr/HanMD15,DBLP:journals/corr/IandolaMAHDK16} and hardware acceleration \cite{fpga,8280474} methods, the intelligent virtual personal assistant has become a hot market (e.g., Siri of iPhone, Alexa of Amazon Echo, Google Now, Samsung Bixby, etc.). 

The most successful deep speech recognition models are built with either CTC \cite{ctc:Graves} or the sequence-to-sequence (seq2seq) model. The CTC based methods \cite{deepspeech2, eesen:Miao} predict a label (i.e., a character label) for each audio frame and is trained to optimize the alignment between the prediction and the target label. Seq2seq models for speech recognition \cite{bahdanau2016end} improved significantly with attention mechanism \cite{attention} which enables more effective flow of information than a recurrent neural network. Attention-based seq2seq model has exceeded the CTC based model on several benchmark datasets. A recent work \cite{joint:Hori} shows that attention-based model may not perform well on noisy speech data, and proposes a joint decoding algorithm for end-to-end ASR with a hybrid CTC/attention architecture. The hybrid method minimizes a joint loss which is the weighted sum of both CTC loss and attention loss, and this effectively utilizes both advantages in decoding. Different from \cite{joint:Hori}, \pname{} uses a cascaded attention-CTC decoder. The benefit of such design includes (1) it avoids tuning the weight parameter in the loss function and (2) it generates the output directly from the softmax output instead of merging the results from both CTC and attention branches.

\subsection{Machine Lipreading}	

Generally, machine lipreading contains two phases: visual feature extraction from the lip movement and text prediction based on classifiers. The traditional visual features extracted from the mouth region of interest (ROI) can be broadly classified into four categories: 1) visual feature extracted directly from images through image transformations such as DCT, DWT \cite{lip:Wu, lip:Morade, lip:Liu, lip:Zhao}; 2) geometry-based features, e.g., height, width, area of the lip region;  3) motion-based features, e.g., optical flow of the motion of the lip region \cite{svm:Shaikh}; 4) model-based features. The shape and appearance of the lip ROI are explicitly modeled by active shape models (ASMs) \cite{lip:Luettin} or active appearance models (AAMs) \cite{lip:Pale, AAM:Cootes}. The quality of such model-based features are highly dependent on the accuracy of hand-labeled training data which requires lots of efforts to obtain. To utilize the temporal dynamics in the extracted visual features, a dynamic classifier such as hidden Markov models (HMMs) is used for classification \cite{lip:Potamianos, lip:Wu}. However, such classifiers are based on the conditional independent assumption and usually fails to model the long-term dependencies.

\begin{figure*}[thpb]
	\centering
	\includegraphics[scale=0.42]{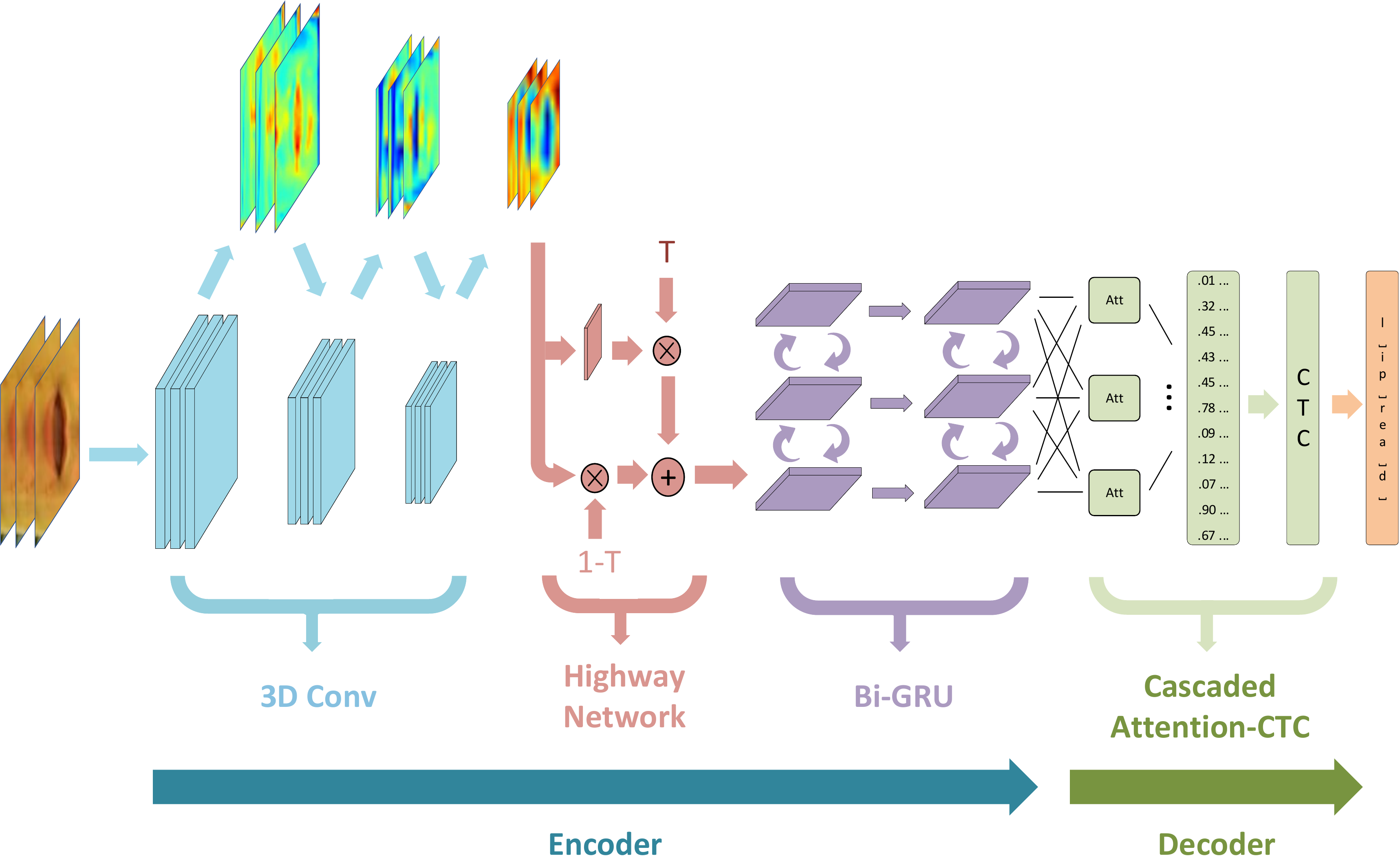}
	\caption{The architecture of \pname{}. The pre-processed mouth-region frames are used as input, which are processed by the following 3D-CNN, the highway network, the Bi-GRU and the cascaded attention-CTC decoder. A softmax layer calculates the probability distribution of the output of the attention module. Each attention unit is related to one character output. The CTC decoder generates text predictions from the probability distribution.}
	\label{fig:ImageEncoder}
\end{figure*}

Recently, deep neural network emerges to be used for lipreading and achieves promising performance compared to the conventional methods. \cite{lipcnn:Noda} propose to use CNN for visual feature extraction, and a Gaussian mixture observation model for an isolated word recognition task by modeling the temporal dependence of the generated phoneme label sequences. 
	\cite{lipnet:Nando} proposes an end-to-end sentence-level lipreading model. A stacked 3D CNN are used for dynamic visual feature extraction. The mapping between lip movement and the corresponding text representation is modeled by the Gated Recurrent Unit (GRU). CTC loss is used for calculating the difference between the predictions and the text labels which have different lengths. 
	\cite{lipnet:Petridis} presents an end-to-end lipreading system which admits two input streams: mouth images that encode static information and mouth differences that capture local temporal dynamics. The encoding layers are pre-trained using RBMs and the temporal dynamics are modeled by an LSTM. The fusion of the two streams is performed through a Bi-directional LSTM (BLSTM).
	\cite{audiovisual:Sanabria, lipwild:Chung} propose to use combined visual speech features for multi-modal speech recognition where video and audio features are fused as input. The former method is based on the `\textit{EESEN}' framework \cite{eesen:Miao}, which uses CTC loss function to solve the temporal alignment problem. The latter approach is based on the `\textit{Watch, Listen, Attend and Spell}' network \cite{las:William}, which leverages the attention mechanism to model rich contextual information between the input video/audio frames and the character predictions.

\section{\pname{} Architecture} \label{sec:proposed}

In this section, we describe the neural network architecture of the proposed \pname{} (Figure~\ref{fig:ImageEncoder}).  The network is composed of a spatio-temporal video encoder which maps a sequence of video frames $\textbf{x} = (x_1, ..., x_n)$  to a sequence of continuous latent representations $\textbf{h} = (h_1, ..., h_n)$, and a cascaded CTC-attention decoder which generates the output sequence of character symbols $\textbf{y} = (y_1, ..., y_m)$\footnote{For lipreading, \textit{m} is normally smaller than \textit{n}, meaning each character symbol corresponds to 1 or more video frames.}.  In the following, we first introduce the spatio-temporal video encoder, and then we present the cascaded attention-CTC decoder.



\subsection{The Spatio-temporal Video Encoder}

The video encoder of \pname{} has three components: \textit{3D-CNN}, \textit{highway network}, and \textit{Bidirectional GRU (Bi-GRU)}.

\textbf{\textit{3D-CNN:}} \pname{} feeds the input video frames into a 3D convolution neural network (3D-CNN) \cite{c3d:Tran} to encode both visual and short temporal information. Compared to 2D-CNN which only deals with the spatial dimensions , 3D-CNN also  convolves across time steps: a group of adjacent \textit{k} frames $x_{t:t+k}$ are processed by a set of \textit{P} temporal convolutional filters $C_p(w, h, k, c)$, where \textit{w} and \textit{h} are the width and height of the filter in the spatial dimension and \textit{c} is the number of channels. The value of \textit{k} in the filter is small (e.g., 3 or 5) to capture only short temporal information, which in lipreading can be mapped to an individual or two nearby visemes.

\begin{figure}[thpb]
	\centering
	\includegraphics[width=0.27\textwidth]{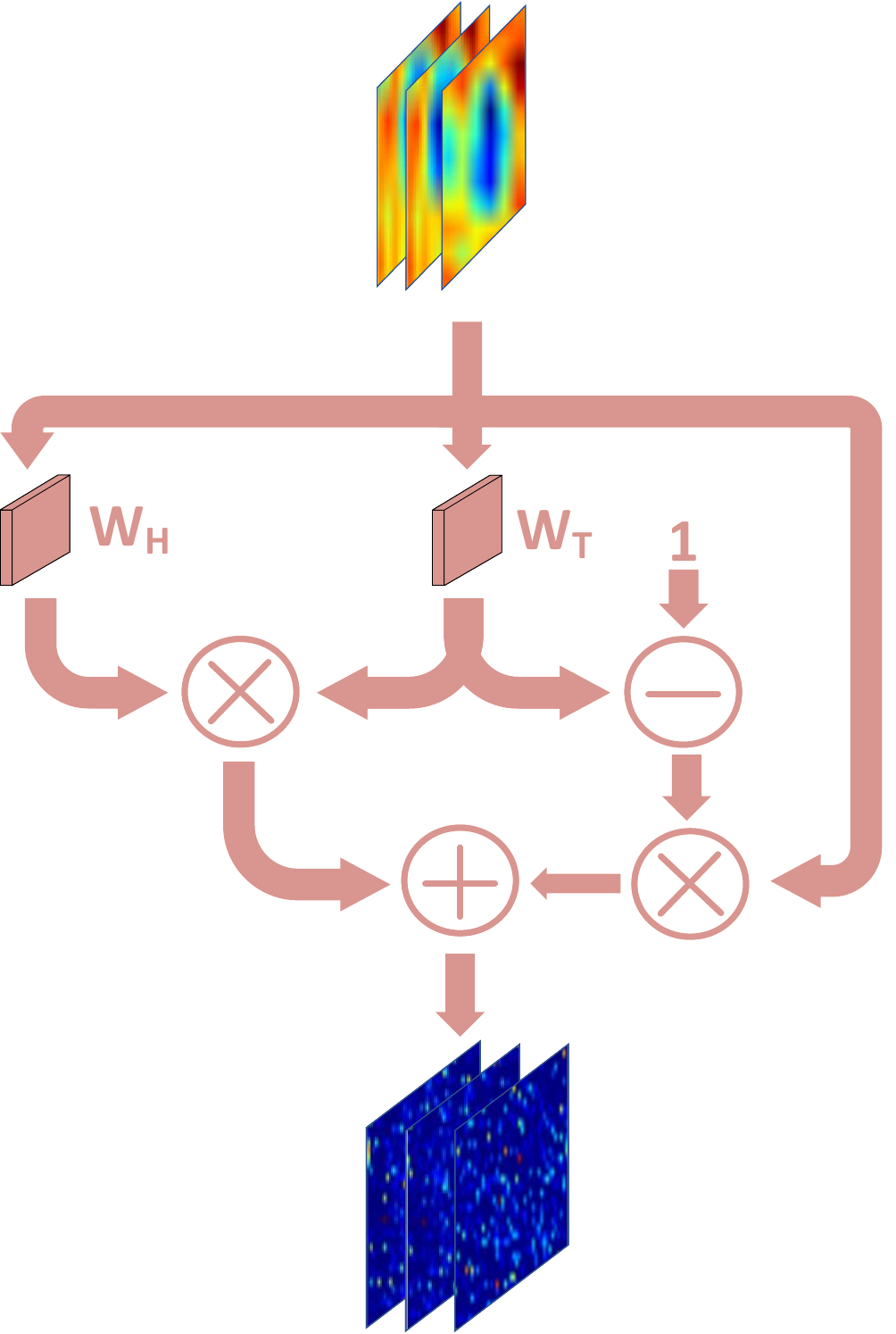}
	\caption{Information flow in the highway network. $W_t$ is the weight of the transform gate and  $1-W_t$ is the weight of the carry gate. The highway network leverages adaptive gating units to regulate the information flow, which allows gradients to flow through many layers.}
	\label{highway}
\end{figure}

\textbf{\textit{Highway Network:}} We stack two layers of highway networks \cite{highway:Srivastava} on top of the 3D-CNN. The highway network module (as shown in Figure \ref{highway}) has a pair of \textit{transform} gate \textit{t} and \textit{carry} gate \textit{1-t} that allows the deep neural network to carry some input information directly to the output. The highway network is formulated as follows:
\begin{align}
\mathbf{t} &= \sigma(\mathbf{W}_T \mathbf{x} + \mathbf{b}_T), \\
\mathbf{g} &= \mathbf{t} \odot \sigma(\mathbf{W}_H \mathbf{x} + \mathbf{b}_H) + (\mathbf{1-t}) \odot \mathbf{x}.
\end{align}
where $\mathbf{x}$ is the input, $\mathbf{g}$ is the network output, and $\odot$ denotes element-wise multiplication. 
A recent study on language modeling \cite{kim2016character} shows that, by adaptively combining local features detected by the individual filters of CNN, highway network enables encoding of much richer semantic features. 
We show that such a design stably enhances the modeling capability of the encoder. An alternative design is to use a 3D CNN with a ResNet, which however highly increases the complexity of the encoder (e.g., a 34-layer ResNet is used in \cite{lipread:Stafylakis}).

\textbf{\textit{Bi-GRU: }} \pname{} encodes long-term temporal information with the Bi-GRU \cite{gru}. 3D-CNN only captures short viseme-level features, however for lipreading problem, there are some visemes visually very similar to each other. Such ambiguity can only be distinguished with longer temporal context which can be well modeled with the state-of-the-art recurrent neural networks such as GRU. A GRU has two gates: 1) a reset gate $r$ which determines how much of the previous memory state should be mixed with the current input; 2) an update gate $z$ which determines how much of the previous memory state should be kept with the current hidden state. The standard formulation is:
\begin{equation}
\begin{aligned}
\mathbf{z}_t &= \sigma(\mathbf{W}_z \mathbf{g}_{t} + \mathbf{U}_z \mathbf{h}_{t-1} + \mathbf{C}_z \mathbf{c}_t + b_z), \\
\mathbf{r}_t &= \sigma(\mathbf{W}_r \mathbf{g}_{t} + \mathbf{U}_r \mathbf{h}_{t-1} + \mathbf{C}_r \mathbf{c}_t + b_r), \\
\tilde{\mathbf{h}}_t  &= \tanh(\mathbf{W}_h \mathbf{g}_{t} + \mathbf{U}_h (\mathbf{r}_t \odot \mathbf{h}_{t-1})  + \mathbf{C}_p \mathbf{c}_t), \\
\mathbf{h}_t &= (1-\mathbf{z}_t) \odot \mathbf{h}_{t-1} + \mathbf{z}_t \odot \tilde{\mathbf{h}}_{t},
\end{aligned}
\end{equation}
where $r_t$ and $z_t$ is the reset and update gate, respectively, $C$ is the linear projection matrix for integrating the context vector, and  $g$ is the output of the highway network. To fully exploit both the past and future context, we use Bi-GRU to capture both the forward and backward information flow. Compared to the vanilla GRU, Bi-GRU can leverage long-range context to improve the motion modeling capacity.


%
%
%

\subsection{Cascaded Attention-CTC Decoder} \label{sec:attention_section}

\subsubsection{Connectionist temporal classification (CTC)}

Many state-of-the-art ASR systems train an end-to-end deep neural network using the Connectionist Temporal Classification (CTC) approach \cite{ctc:Graves}. CTC is an objective loss function that eliminates the requirement of explicit alignment between the input sequence and the target sequence during the model training process.  Given an input $\mathbf{x}$ of length $T$, the output of the neural network is a softmax layer which indicates the possibility of emitting the symbol (either a character or $`blank'$) at every time-step. For the same target sequence, there can be many alignment candidates since the blank symbol can be inserted at any position. For example, the paths (x, -, -, y, -, z), (-, x, -, y, -, z) and (-, x, -, y, z, -) all map to the same target label of (x, y, z).  Now, the possibility of generating one correct path $\pi$ for the given input $\mathbf{x}$ is: 
\begin{equation}\label{ctc_path}
p(\pi|\mathbf{x}) = \prod_{t=1}^{T}p(\pi_t|\mathbf{x}).
\end{equation}
The probability of generating the target sequence $\mathbf{l}$ is the sums over all possible paths $\pi$:
\begin{equation}\label{ctc_prob}
p(\text{l} |\mathbf{x}) = \sum_{i} p(\pi_i | \mathbf{x}).
\end{equation}
A forward-backward dynamic programming algorithm is proposed to calculate equation (\ref{ctc_prob}) efficiently and makes the network training possible. 
Finally, the CTC network is trained to minimize the negative log-likelihood of the target sequence $\mathbf{l}$ as:
\begin{equation}\label{ctc_loss}
\mathcal{L}_{CTC} =  - \ln p(\text{l} |\mathbf{x}).
\end{equation}

According to equation \ref{ctc_path}, the CTC approach assumes conditional independence of separate labels (i.e., the character symbols predicted by $p(\pi_t|\mathbf{x})$), and thus obstructing further performance improvement.

\subsubsection{Cascaded Attention-CTC}



\begin{figure}[thpb]
	\centering
	\includegraphics[width=0.24\textwidth]{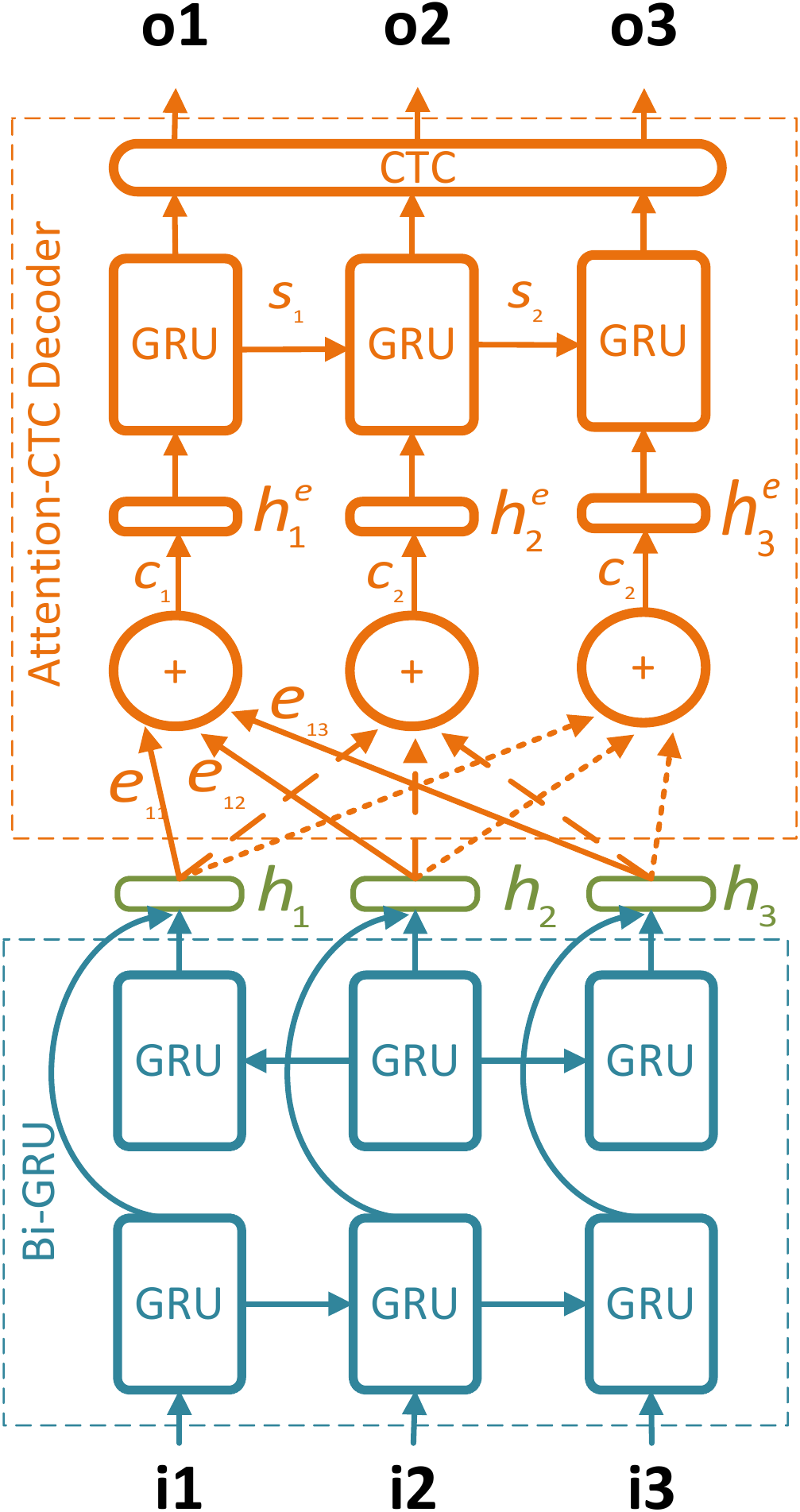}
	\caption{Diagram of the proposed cascaded attention-CTC decoder. $i$ and $o$ are the input and output sequence, respectively. $h_i$ is the intermediate hidden state of the Bi-GRU. $e_{i,j}$ is the weight of the attention unit, where $i, j$ denotes the index of the output and input node, respectively. Each attention unit generate a fixed-lenght context vector based on the input sequence.}
	\label{fig:attention}
\end{figure}

To capture information explicitly from longer context, \pname{} feeds the encoded spatiotemperal features into the cascaded attention-CTC decoder (Figure \ref{fig:attention}). As illustrated here, the decoder uses an attention mechanism to find an alignment between each element of the decoded sequence ($\mathbf{s}_1, \mathbf{s}_2, \mathbf{s}_3$) and the hidden states from the encoder ($\mathbf{h}_1, \mathbf{h}_2, \mathbf{h}_3$). The attention score is calculated as:
\begin{align}
e_{j,t} &= \mathbf{V}_a \tanh(\mathbf{W}_a \mathbf{s}_{t-1} + \mathbf{U}_a \mathbf{h}_j), \\
\alpha_{j,t} &= \frac{\exp(e_j)}{\sum_{k=1}^{T}\exp(e_k)} .
\end{align}

The context vector is computed as the weighted average over all the source hidden states.
\begin{equation}
c_t = \sum_{k=1}^{T} \alpha_{k,t} h_k
\end{equation}

The prediction result for the frame at time step \textit{t} is computed based on the context, hidden state, and the previous prediction:
\begin{equation}
y_t = \sigma(\mathbf{W}_o \mathbf{E} \mathbf{y}_{t-1} + \mathbf{U}_o \mathbf{h}_{t-1} + \mathbf{C}_o \mathbf{c}_t).
\end{equation}
where $\mathbf{E}$ is the character-level embedding matrix which is much simpler than the word embedding matrix used and learned in neural machine translation tasks. The output prediction is fed into the CTC loss function to guide the model training process.




Attention mechanism debilitates the constraint of the conditional independence assumption in CTC loss. Therefore it improves the modeling capability on the lipreading problem and can give better predictions on visually similar visemes. On the other hand, compared to the attention only architecture, the combination of attention mechanism and CTC substantially reduces the irregular alignments during the training stage, as CTC will guide the attention decoder to eliminate the unnecessary non-sequential predictions between the hypothesis and reference. Therefore, the attention module in LCANet requires fewer glimpses to find the correct focus-then-decode pattern. From the experimental result, the cascaded attention-CTC decoder outperforms existing methods. 



    \section{Experiments}
In this section, we evaluate the performance of the proposed \pname{} and compare it with other state-of-the-art lipreading methods on public benchmark datasets.

\subsection{Datasets}
\textbf{GRID} The GRID dataset contains 28 hours of videos which are recorded from 34 different speakers each speaks 1000 sentences in a controlled lab environment. Each sentence has a fixed 6-word structure: $\textit{command + color + preposition + letter + digit + adverb}$, e.g., $\textit{``set blue with h seven again"}$. There are 51 different words including four commands, four colors, four prepositions, 25 letters, ten digits and four adverbs, yielding 64,000 possible sentences. For each sentence, the six words are randomly combined. Until the paper submission, GRID is still the largest available public sentence level lip-reading dataset. We follow the evaluation protocol of \cite{lipnet:Nando, lipwild:Chung} to randomly divide the dataset into train, validation and test sets, where 255 sentences of each speaker are used for testing. All videos are recorded with a frame rate of 25fps and each lasts for 3 seconds (i.e., 75 frames per sample). Based on 68 facial landmarks extracted by Dlib \cite{dlib:Davis}, we crop 100x50 affine-transformed mouth-centered region as the input data. We then apply an affine transformation to extract an aligned mouth region and each has a resolution of $100 \times 50$. The obtained RGB mouth images are mean-reduced and normalized. We perform data augmentation on the training data similar to that used in \cite{lipnet:Nando} to avoid overfitting. 


\subsection{Evaluation Metrics}

We use character error rate (CER), word error rate (WER), and BLEU \cite{papineni2002bleu} to benchmark the proposed framework. CER is computed with the minimum number of operations required to transform the predicted text to the ground truth (e.g., the Levenshtein distance), divided by the number of characters in the ground truth. WER is similar to CER except it operates on a word level. Smaller CER/WER means higher prediction accuracy, while a larger BLEU score is preferred. 

\subsection{Implementation Details}

We implement and train \pname{} using Tensorflow \cite{tensorflow:Martin}. The detailed parameters of each layer in the architecture are summarized in Table~\ref{tab:layer}. ADAM optimizer \cite{adam:Kingma} is used with an initial learning rate of 0.0001 and the batch size is 64. We train the propose model 50 epochs on GRID with early stopping. 

\begin{table}[]
	\centering
	\caption{The detailed architecture of the proposed method.}
	\label{tab:layer}
	\resizebox{0.48\textwidth}{!}{
	\begin{tabular}{|c|c|c|c|c|}
		\hline
		Layer name                         & Output size                      & Kernel                & Stride                & Pad                   \\ \hline
		3d-conv1                           & 75x50x25x32                      & 3, 5, 5               & 1, 2, 2               & 1, 2, 2               \\ 
		\multicolumn{1}{|l|}{bn/relu/drop} & \multicolumn{1}{l|}{75x50x25x32} & \multicolumn{1}{l|}{} & \multicolumn{1}{l|}{} & \multicolumn{1}{l|}{} \\ 
		pool1                              & 75x25x12x32                      & 1, 2, 2               & 1, 2, 2               &                       \\ \hline
		3d-conv2                           & 75x25x12x64                      & 3, 5, 5               & 1, 1, 1               & 1, 2, 2               \\ 
		\multicolumn{1}{|l|}{bn/relu/drop} & \multicolumn{1}{l|}{75x25x12x64} & \multicolumn{1}{l|}{} & \multicolumn{1}{l|}{} & \multicolumn{1}{l|}{} \\ 
		pool2                              & 75x12x6x64                       & 1, 2, 2               & 1, 2, 2               &                       \\ \hline
		3d-conv3                           & 75x12x6x96                       & 3, 5, 5               & 1, 2, 2               & 1, 2, 2               \\ 
		\multicolumn{1}{|l|}{bn/relu/drop} & \multicolumn{1}{l|}{75x12x6x96}  & \multicolumn{1}{l|}{} & \multicolumn{1}{l|}{} & \multicolumn{1}{l|}{} \\ 
		pool3                              & 75x6x3x96                        & 1, 2, 2               & 1, 2, 2               &                       \\ \hline
		highway1                           & 75x1728                          &                       &                       &                       \\ 
		highway2                           & 75x1728                          &                       &                       &                       \\ \hline
		gru1                               & 75x512                           &                       &                       &                       \\ 
		gru2                               & 75x512                           &                       &                       &                       \\ \hline
		atten1                             & 75x28                            &                       &                       &                       \\ \hline
		CTC Loss                           &     75                             &                       &                       &                       \\ \hline
	\end{tabular}
}
\end{table}

\begin{figure}%
	\centering
	\subfloat[\label{fig:AH-CTC}]{\includegraphics[width=.32\linewidth]{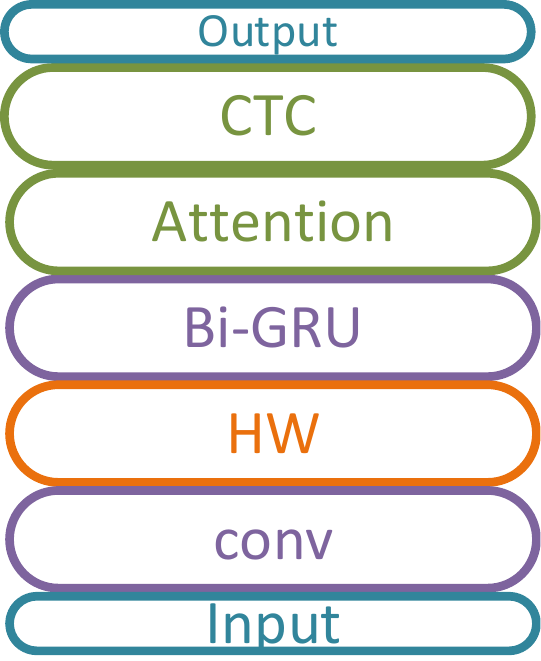}}
	\hspace{10pt}%
	\subfloat[\label{fig:AH-CTC-CE}]{\includegraphics[width=.32\linewidth]{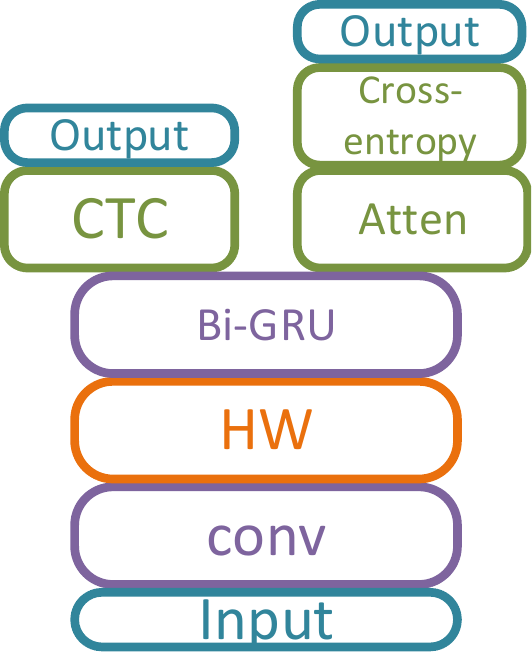}} \\
	\hspace{10pt}%
	\subfloat[\label{fig:A-CTC}]{\includegraphics[width=.33\linewidth]{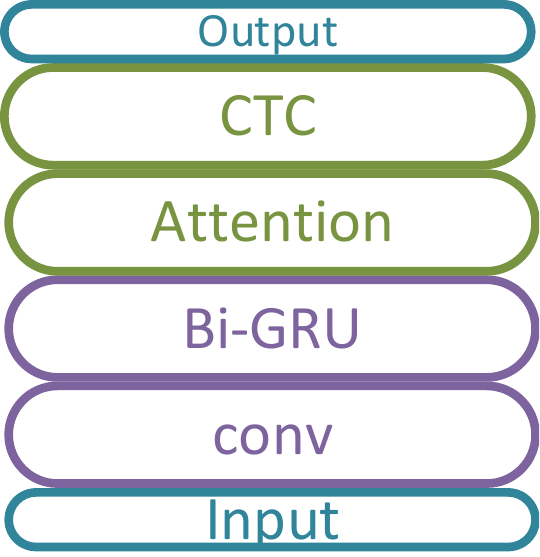}} 
	\hspace{10pt}%
	\subfloat[\label{fig:H-CTC}]{\includegraphics[width=.32\linewidth]{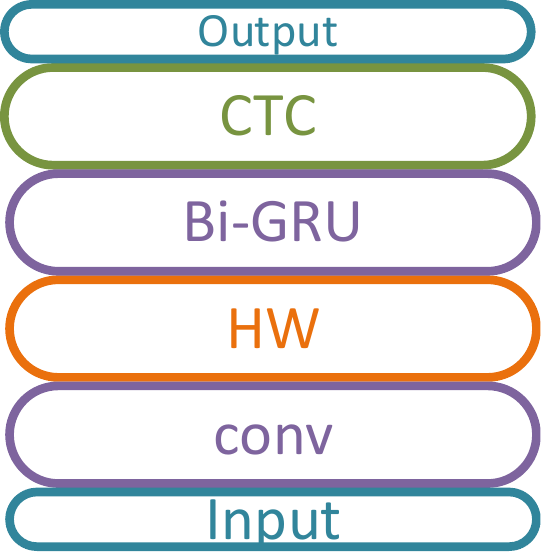}} 
	\caption{Diagram of the variant models for comparison. (a) AH-CTC (\pname{}), (b) AH-CTC-CE, (c) A-CTC,  (d) H-CTC.}%
	\label{fig:implementation}%
\end{figure}

The proposed approach is compared with \cite{lipnet:Nando, lipread:Chung, lipwild:Chung}, which is referred as `Lipnet', `LRSW' and `LRW', respectively. LipNet is the first end-to-end sentence-level lipreading model. LRW and LRSW achieve state-of-the-art performance on the ``Lip Reading Sentences" dataset and ``Lip Reading in the wild" dataset, respectively. In addition, we compare \pname{} with several of its variances to explore the impact of the highway network and the cascaded attention-CTC decoder. The structures of these models are described in Table~\ref{tab:implementation} and the simplified diagrams are shown in Figure~\ref{fig:implementation}. The model architectures in subfigures~\ref{fig:AH-CTC}, \ref{fig:A-CTC} and \ref{fig:H-CTC} all use a single CTC loss but varies on whether the highway network layers or the attention mechanism is used. We also compare a multi-task learning architecture \cite{joint:Hori} (subfigure~\ref{fig:AH-CTC-CE}) which optimizes a joint loss in two branches: CTC processes one branch, and the joint attention/cross-entropy loss handle another. The objective loss function is defined as:
\begin{equation}
	\mathcal{L}_{total} = \lambda * \mathcal{L}_{ctc} + (1-\lambda) * \mathcal{L}_{atten}.
\end{equation}
The final prediction result is generated with a joint decoder as described in \cite{joint:Hori}.
In the experiment, we tried different $\lambda$ values (from 0.1 to 0.9) and only report the best result here.




\begin{table}[]
	\centering
		\caption{Structure description of the proposed method and its variants. AH-CTC, H-CTC, and A-CTC use only the CTC loss function. While the AH-CTC-CE model leverages the cross-entropy loss on the attention branch and the CTC loss on another branch.}
	\label{tab:implementation}
	\resizebox{0.48\textwidth}{!}{
	\begin{tabular}{|c|c|c|c|c|}
		\hline
		\multirow{2}{*}{Method} & \multicolumn{4}{c|}{Structure}                                                               \\ \cline{2-5} 
				& Attention & \begin{tabular}[c]{@{}c@{}}Highway \\ network\end{tabular} & CTC loss & Cross-entropy loss\\ \hline

		AH-CTC-CE    &    \checkmark       &       \checkmark        &         \checkmark &          \checkmark                \\
		H-CTC    &           &          \checkmark       &            \checkmark          &                                    \\ 
		A-CTC    &    \checkmark       &                     &      \checkmark                &                  \\ \hline 
		\textbf{AH-CTC}$(\textbf{\pname{}})$                  &      \checkmark     &        \checkmark              &          \checkmark            &           \\ \hline
	\end{tabular}
}
\end{table}

\subsection{Results}
The performance comparison between the proposed method and the baselines on the GRID dataset is summarized in Table~\ref{tab:grid}. Our architecture performs character-level prediction on input frames. A word is regarded as correctly predicted when "every" character is correct.  \pname{} achieves 1.3\% CER and 2.9\% WER.  Compared to Lipnet, \pname{} has 31.6$\%$ and 37.5$\%$ improvement on CER and WER, receptively. The LRSW model is pre-trained on the large-scale ``Lip Reading Sentences" dataset proposed in \cite{lipwild:Chung}. However, our method still achieves better CER and WER. 

The effectiveness of the highway network layers and the cascaded attention-CTC decoder is demonstrated by the comparison among models $\{3, 4, 5, 6\}$ in Table~\ref{tab:grid}. First, we show that the \pname{} (model $\{6\}$) is better than the multi-task or multi-branch variant (model $\{3\}$). The reason is partially because only the shared encoder is jointly updated by both the attention and CTC module in the multi-task model. Also, since the multi-task model relies on the CTC loss and the cross-entropy loss, it is difficult to coordinate the behavior of the two modules. In contrast, the proposed cascaded architecture uses a single CTC loss to update both the encoder and decoder. The leverage of the attention mechanism guarantees output sequences generated from input sources with long-term coherence. The CTC decoder enforces monotonic alignment to guide the attention module to concentrate on the salient sequential parts of the input video frames. Second, the benefit of the highway network is illustrated by comparing \pname{} with model $\{5\}$, which yields 0.4\% improvement on CER and 1.1\% improvement on WER. The function of the highway network is similar to the ResNet \cite{resnet:He}. The difference is the highway network leverages adaptive gating units to regulate the information flow, which allows gradients to flow through many layers without attenuation. Finally, removing the attention mechanism (model $\{4\}$) also greatly compromises the overall performance. In addition, model $\{5\}$ can be regarded as the attention-only model. The comparison between model $\{5\}$ and model $\{6\}$ illustrates the proposed cascaded attention-CTC model is better than its attention-only variant.





\begin{table}[h]
	\centering
	\caption{Performance on the GRID dataset.}
	\label{tab:grid}
	\resizebox{0.48\textwidth}{!}{
		\begin{tabular}{|c|c|c|c|c|}
			\hline
			$\#$  &  Method                   & CER $\%$ & WER $\%$ & BLEU $\%$\\ \hline
			1 & Lipnet \cite{lipnet:Nando}                   & 1.9 & 4.8  & 96.0 \\ 
			2 & LRSW \cite{lipwild:Chung}                      &     & 3.0  &      \\ \hline
			3 & AH-CTC-CE                & 2.1 & 4.9  & 95.8 \\ 
			4 & H-CTC                    & 1.8 & 4.3  & 96.4 \\ 
			5 & A-CTC                    & 1.7 & 4.1 & 96.7 \\ \hline
			6 & \textbf{AH-CTC}$(\textbf{\pname{}})$ & \textbf{1.3} & \textbf{2.9}  & \textbf{97.4} \\ \hline
		\end{tabular}
	}
\end{table}



\subsection{Convergence speed}
\begin{figure}[thpb]
	\centering
	\includegraphics[width=0.44\textwidth]{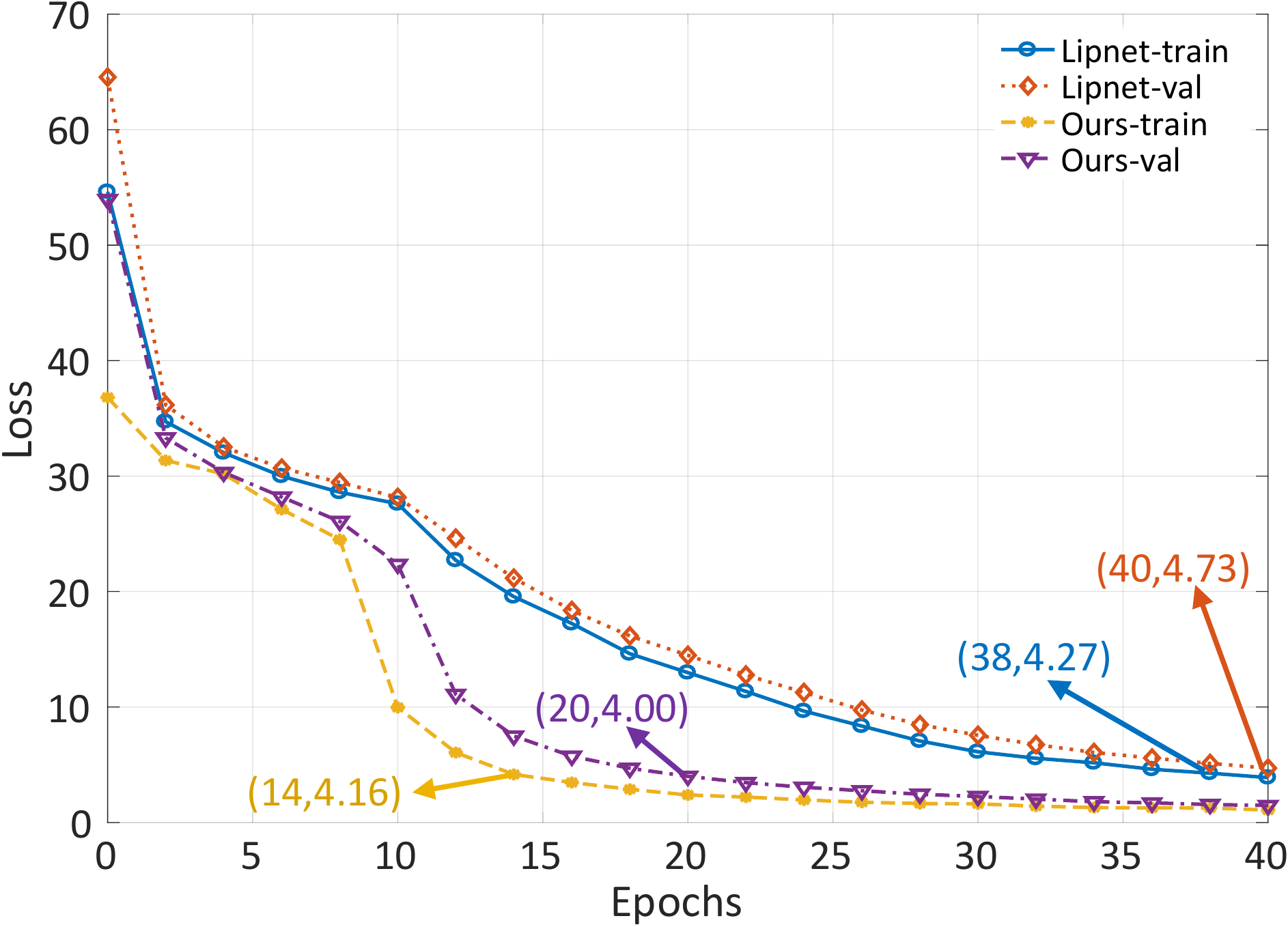}
	\caption{Training curve. The value pair ($\#$iteration, loss) in the figure denotes the required iterations to achieve the specified loss.}
	\label{fig:lr}
\end{figure}

The training and validation loss on GRID dataset for the proposed approach and Lipnet are shown in Figure~\ref{fig:lr}. In the figure, we mark the required number of epochs for the loss to reduce to 4. \pname{} reaches the loss value 4 in 14 epochs for training and 20 epochs for validation. On the contrary, the converge is much slower for Lipnet, which reaches the same loss in 38 epochs for training and 40 epochs for validation. The speed-up comes from the cascaded attention-CTC module. The attention mechanism generates output sequences with long-term temporal correlation, which overcomes the disadvantage of CTC that assumes inputs are conditionally independent. Besides, the attention mechanism can be treated as a pre-alignment operation that eliminates unnecessary paths for the CTC decoder. Meanwhile, CTC guarantees a monotonic alignment between video frames and text labels, which helps to guide the attention mechanism to focus on the sequential order video-text pairs. Therefore, less irrelevant samples will be generated that leads to the observed speed-up. 

\subsection{Confusion Matrix} 
\begin{figure}%
		\centering
		\subfloat[\label{fig:lipround}]{\includegraphics[width=0.9\linewidth]{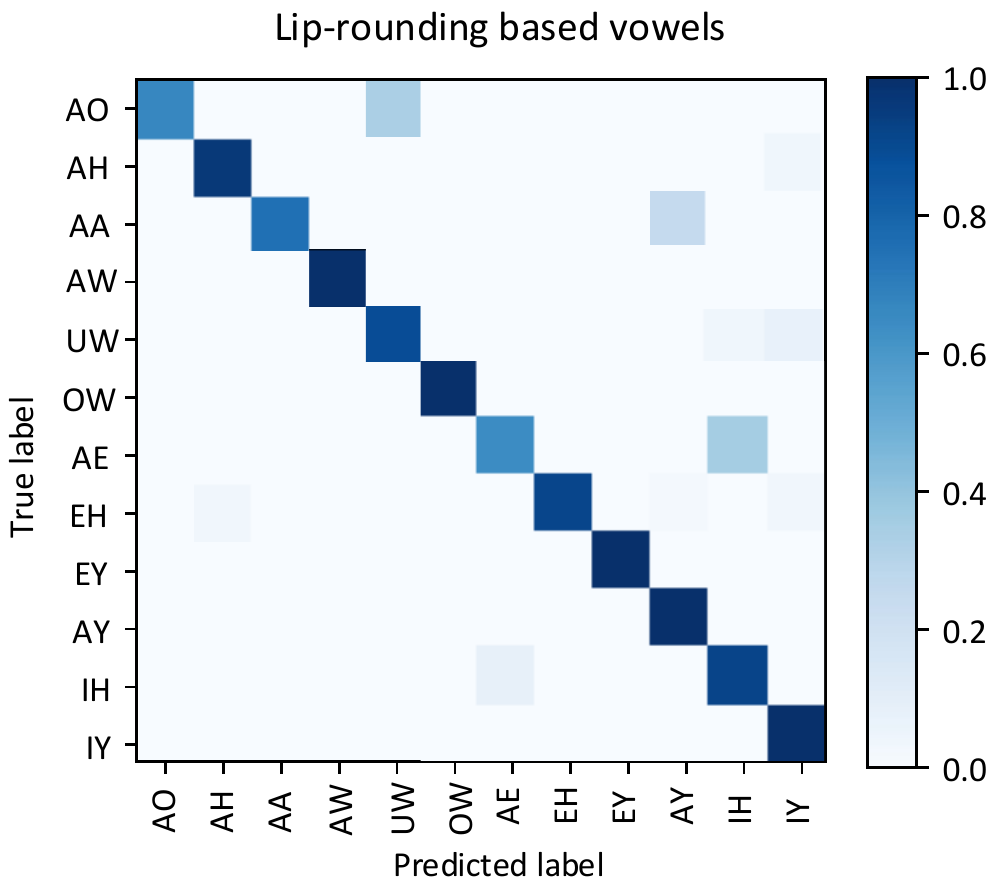}} \\
		\vspace{-10pt}
		\subfloat[\label{fig:bilabial}]{\includegraphics[width=.45\linewidth]{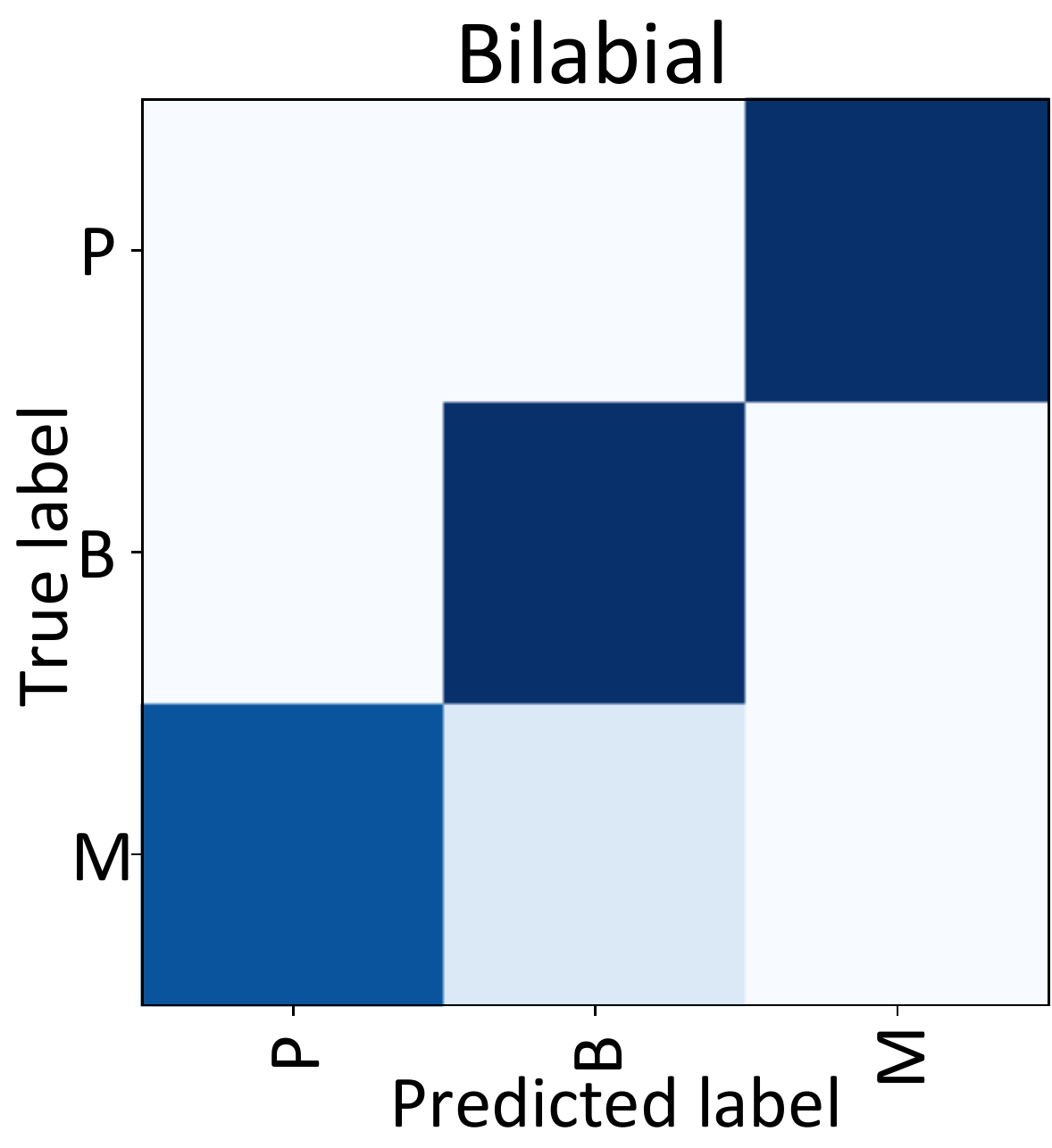}} 
		\subfloat[\label{fig:Intra-visemes}]{\includegraphics[width=.45\linewidth]{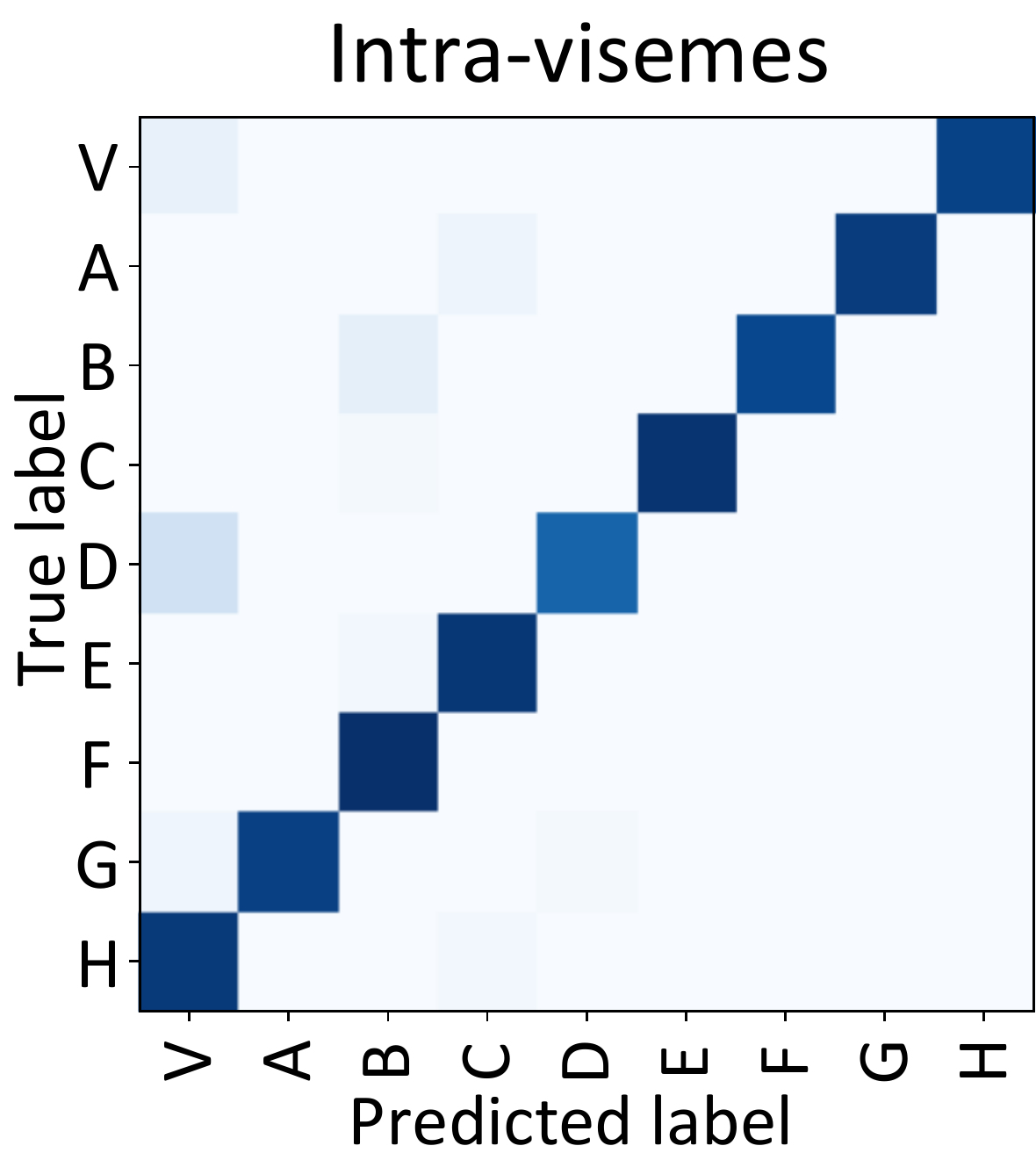}} 
		\caption{(a) Lip-rounding based vowels, (b) Bilabial , (c) Intra-visemes.}%
		\label{fig:phoneme}%
	\end{figure}

	We use the mapping proposed by the IBM ViaVoice database \cite{mapping:Neti}. 43 phonemes are grouped into 13 visemes classes (including a silence class), denoted by: lip-rounding based vowels (V), Alveolar-semivowels (A), Alveolar-fricatives (B), Alveolar (C), Palato-alveolar (D), Bilabial (E), Dental (F), Labio-dental (G), Velar (H). We plot the confusion matrix for the viseme class of lip-rounding based vowels, Bilabial in Figure~\ref{fig:phoneme} since they capture the most confusing phonemes. For example, in Figure~\ref{fig:lipround}, $\{/AA/, /AY/\}$, $\{/AE/, /IH/\}$ and $\{/AO/, /UW/\}$ are frequently incorrectly classified during the text decoding process. This phenomenon is caused by the similar pronunciation between the text pair of $\{r, i\}$, $\{at, bin\}$, $\{four, two\}$. The intra-visemes categorical confusion matrix is shown in Figure~\ref{fig:Intra-visemes}. The misclassification between different viseme categories is not statistically significant. There is a tiny uncertainty between viseme class V and D, which demonstrates the proposed method is effective on the lipreading task. 

	
    \section{Conclusions}
	In this paper, we have presented \pname{}, an end-to-end deep neural network architecture for machine lipreading. To accurately predict visemes, \pname{} introduced a cascaded attention-CTC decoder which effectively compensates the defect of the conditional independence assumption of the CTC approach. In addition, \pname{} stacks highway network layers over 3D-CNN layers to further improve performance. Extensive experiment results show that \pname{} achieves the state-of-the-art accuracy as well as faster convergence. 
	
	Our future work lies in two directions. First, \pname{} currently deals with the conditional independence assumption of CTC using attention mechanisms, but the loss function itself is not modified. Therefore, we will investigate solutions that can be directly applied on the loss function. Second, we would like to extend \pname{} with joint audio and visual input as \cite{lipwild:Chung} to explore further performance improvement.

%
%
	
	\bibliographystyle{IEEEtran}
	\bibliography{ref}

\begin{thebibliography}{10}
\providecommand{\url}[1]{#1}
\csname url@samestyle\endcsname
\providecommand{\newblock}{\relax}
\providecommand{\bibinfo}[2]{#2}
\providecommand{\BIBentrySTDinterwordspacing}{\spaceskip=0pt\relax}
\providecommand{\BIBentryALTinterwordstretchfactor}{4}
\providecommand{\BIBentryALTinterwordspacing}{\spaceskip=\fontdimen2\font plus
\BIBentryALTinterwordstretchfactor\fontdimen3\font minus
  \fontdimen4\font\relax}
\providecommand{\BIBforeignlanguage}[2]{{%
\expandafter\ifx\csname l@#1\endcsname\relax
\typeout{** WARNING: IEEEtran.bst: No hyphenation pattern has been}%
\typeout{** loaded for the language `#1'. Using the pattern for}%
\typeout{** the default language instead.}%
\else
\language=\csname l@#1\endcsname
\fi
#2}}
\providecommand{\BIBdecl}{\relax}
\BIBdecl

\bibitem{hilder2009comparison}
S.~Hilder \emph{et~al.}, ``Comparison of human and machine-based lip-reading,''
  in \emph{INTERSPEECH}, 2009.

\bibitem{lipnet:Petridis}
S.~Petridis \emph{et~al.}, ``End-to-end visual speech recognition with lstms,''
  in \emph{ICASSP}, 2017.

\bibitem{lipnet:Nando}
Y.~M. Assael \emph{et~al.}, ``Lipnet: Sentence-level lipreading,'' \emph{CoRR},
  vol. abs/1611.01599, 2016.

\bibitem{lipreading:Wand}
M.~Wand \emph{et~al.}, ``Lipreading with long short-term memory,'' in
  \emph{ICASSP}, 2016.

\bibitem{lipwild:Chung}
J.~S. Chung \emph{et~al.}, ``Lip reading sentences in the wild,'' in
  \emph{CVPR}, 2017.

\bibitem{lipcnn:Noda}
K.~Noda \emph{et~al.}, ``Lipreading using convolutional neural network,'' in
  \emph{INTERSPEECH}, 2014.

\bibitem{geng2017novel}
Y.~Geng, G.~Zhang, W.~Li, Y.~Gu, R.-Z. Liang, G.~Liang, J.~Wang, Y.~Wu,
  N.~Patil, and J.-Y. Wang, ``A novel image tag completion method based on
  convolutional neural transformation,'' in \emph{International Conference on
  Artificial Neural Networks}.\hskip 1em plus 0.5em minus 0.4em\relax Springer,
  2017, pp. 539--546.

\bibitem{zhang2017learning}
G.~Zhang, G.~Liang, W.~Li, J.~Fang, J.~Wang, Y.~Geng, and J.-Y. Wang,
  ``Learning convolutional ranking-score function by query preference
  regularization,'' in \emph{International Conference on Intelligent Data
  Engineering and Automated Learning}.\hskip 1em plus 0.5em minus 0.4em\relax
  Springer, 2017, pp. 1--8.

\bibitem{miao2016cnn}
S.~Miao, Z.~J. Wang, and R.~Liao, ``A cnn regression approach for real-time
  2d/3d registration,'' \emph{IEEE transactions on medical imaging}, vol.~35,
  no.~5, pp. 1352--1363, 2016.

\bibitem{Tome_2017_CVPR}
D.~Tome, C.~Russell, and L.~Agapito, ``Lifting from the deep: Convolutional 3d
  pose estimation from a single image,'' in \emph{The IEEE Conference on
  Computer Vision and Pattern Recognition (CVPR)}, July 2017.

\bibitem{peng2017reconstruction}
X.~Peng, X.~Yu, K.~Sohn, D.~Metaxas, and M.~Chandraker, ``Reconstruction for
  feature disentanglement in pose-invariant face recognition,'' \emph{arXiv
  preprint arXiv:1702.03041}, 2017.

\bibitem{wu2015fully}
J.~Wu, E.~E. Abdel-Fatah, and M.~R. Mahfouz, ``Fully automatic initialization
  of two-dimensional--three-dimensional medical image registration using hybrid
  classifier,'' \emph{Journal of Medical Imaging}, vol.~2, no.~2, p. 024007,
  2015.

\bibitem{wu2016robust}
J.~Wu and M.~R. Mahfouz, ``Robust x-ray image segmentation by spectral
  clustering and active shape model,'' \emph{Journal of Medical Imaging},
  vol.~3, no.~3, p. 034005, 2016.

\bibitem{ding2016articulated}
M.~Ding and G.~Fan, ``Articulated and generalized gaussian kernel correlation
  for human pose estimation,'' \emph{IEEE Transactions on Image Processing},
  vol.~25, no.~2, pp. 776--789, 2016.

\bibitem{meng_Face}
J.~Liang, C.~Chen, Y.~Yi, X.~Xu, and M.~Ding, ``Bilateral two-dimensional
  neighborhood preserving discriminant embedding for face recognition,''
  \emph{IEEE Access}, vol.~5, 2017.

\bibitem{wang2013automatic}
X.~Wang, L.~Pollock, and K.~Vijay-Shanker, ``Automatic segmentation of method
  code into meaningful blocks: Design and evaluation,'' \emph{Journal of
  Software: Evolution and Process}, 2013.

\bibitem{7884622}
------, ``Automatically generating natural language descriptions for
  object-related statement sequences,'' in \emph{Proceedings of the 24th
  International Conference on Software Analysis, Evolution and Reengineering
  (SANER)}, Feb 2017, pp. 205--216.

\bibitem{deepspeech2}
D.~Amodei \emph{et~al.}, ``Deep speech 2: End-to-end speech recognition in
  english and mandarin,'' in \emph{ICML}, 2016.

\bibitem{eesen:Miao}
Y.~Miao \emph{et~al.}, ``Eesen: End-to-end speech recognition using deep rnn
  models and wfst-based decoding.'' in \emph{In IEEE Workshop on Automatic
  Speech Recognition and Understanding (ASRU)}, 2015.

\bibitem{asr:Zhang}
Y.~{Zhang} \emph{et~al.}, ``{Towards End-to-End Speech Recognition with Deep
  Convolutional Neural Networks},'' \emph{arXiv:1701.02720}, 2017.

\bibitem{attention}
D.~{Bahdanau} \emph{et~al.}, ``{Neural Machine Translation by Jointly Learning
  to Align and Translate},'' \emph{arXiv:1409.0473}, 2014.

\bibitem{sennrich2015neural}
R.~Sennrich \emph{et~al.}, ``Neural machine translation of rare words with
  subword units,'' \emph{arXiv:1508.07909}, 2015.

\bibitem{gehring2017convolutional}
J.~Gehring \emph{et~al.}, ``Convolutional sequence to sequence learning,''
  \emph{aarXiv:1705.03122}, 2017.

\bibitem{vaswani2017attention}
A.~Vaswani \emph{et~al.}, ``Attention is all you need,''
  \emph{arXiv:1706.03762}, 2017.

\bibitem{ctc:Graves}
A.~Graves \emph{et~al.}, ``Connectionist temporal classification: Labelling
  unsegmented sequence data with recurrent neural networks,'' in \emph{ICML},
  2006.

\bibitem{GRID}
C.~Chandrasekaran \emph{et~al.}, ``The natural statistics of audiovisual
  speech,'' \emph{PLoS computational biology}, vol.~5, no.~7, p. e1000436,
  2009.

\bibitem{chen2017progressive}
Z.~Chen \emph{et~al.}, ``Progressive joint modeling in unsupervised
  single-channel overlapped speech recognition,'' \emph{arXiv:1707.07048},
  2017.

\bibitem{erdogan2016multi}
H.~Erdogan \emph{et~al.}, ``Multi-channel speech recognition: Lstms all the way
  through,'' in \emph{CHiME-4 workshop}, 2016.

\bibitem{dahl2012context}
G.~E. Dahl \emph{et~al.}, ``Context-dependent pre-trained deep neural networks
  for large-vocabulary speech recognition,'' \emph{IEEE Transactions on audio,
  speech, and language processing}, vol.~20, no.~1, pp. 30--42, 2012.

\bibitem{asr:Graves}
A.~Graves \emph{et~al.}, ``Speech recognition with deep recurrent neural
  networks,'' in \emph{ICASSP}, 2013.

\bibitem{hinton2012deep}
G.~Hinton \emph{et~al.}, ``Deep neural networks for acoustic modeling in speech
  recognition: The shared views of four research groups,'' \emph{IEEE Signal
  Processing Magazine}, vol.~29, no.~6, pp. 82--97, 2012.

\bibitem{bahdanau2016end}
D.~Bahdanau \emph{et~al.}, ``End-to-end attention-based large vocabulary speech
  recognition,'' in \emph{ICASSP}, 2016.

\bibitem{las:William}
W.~Chan \emph{et~al.}, ``Listen, attend and spell: A neural network for large
  vocabulary conversational speech recognition,'' in \emph{ICASSP}, 2016.

\bibitem{kim2016character}
Y.~Kim \emph{et~al.}, ``Character-aware neural language models,'' in
  \emph{AAAI}, 2016.

\bibitem{nlp2}
A.~Kumar \emph{et~al.}, ``Ask me anything: Dynamic memory networks for natural
  language processing,'' in \emph{ICML}, 2016.

\bibitem{nlp1}
A.~Conneau \emph{et~al.}, ``Very deep convolutional networks for natural
  language processing,'' \emph{arXiv:1606.01781}, 2016.

\bibitem{nlp3}
J.~Weston \emph{et~al.}, ``Towards ai-complete question answering: A set of
  prerequisite toy tasks,'' \emph{arXiv:1502.05698}, 2015.

\bibitem{deeprebirth}
\BIBentryALTinterwordspacing
D.~Li, X.~Wang, and D.~Kong, ``Deeprebirth: Accelerating deep neural network
  execution on mobile devices,'' \emph{CoRR}, vol. abs/1708.04728, 2017.
  [Online]. Available: \url{http://arxiv.org/abs/1708.04728}
\BIBentrySTDinterwordspacing

\bibitem{DBLP:journals/corr/HanMD15}
\BIBentryALTinterwordspacing
S.~Han, H.~Mao, and W.~J. Dally, ``Deep compression: Compressing deep neural
  network with pruning, trained quantization and huffman coding,'' \emph{CoRR},
  vol. abs/1510.00149, 2015. [Online]. Available:
  \url{http://arxiv.org/abs/1510.00149}
\BIBentrySTDinterwordspacing

\bibitem{DBLP:journals/corr/IandolaMAHDK16}
\BIBentryALTinterwordspacing
F.~N. Iandola, M.~W. Moskewicz, K.~Ashraf, S.~Han, W.~J. Dally, and K.~Keutzer,
  ``Squeezenet: Alexnet-level accuracy with 50x fewer parameters and
  {\textless}1mb model size,'' \emph{CoRR}, vol. abs/1602.07360, 2016.
  [Online]. Available: \url{http://arxiv.org/abs/1602.07360}
\BIBentrySTDinterwordspacing

\bibitem{fpga}
\BIBentryALTinterwordspacing
S.~Han, J.~Kang, H.~Mao, Y.~Hu, X.~Li, Y.~Li, D.~Xie, H.~Luo, S.~Yao, Y.~Wang,
  H.~Yang, and W.~B.~J. Dally, ``Ese: Efficient speech recognition engine with
  sparse lstm on fpga,'' in \emph{Proceedings of the 2017 ACM/SIGDA
  International Symposium on Field-Programmable Gate Arrays}, ser. FPGA
  '17.\hskip 1em plus 0.5em minus 0.4em\relax New York, NY, USA: ACM, 2017, pp.
  75--84. [Online]. Available: \url{http://doi.acm.org/10.1145/3020078.3021745}
\BIBentrySTDinterwordspacing

\bibitem{8280474}
Y.~Mao, J.~Oak, A.~Pompili, D.~Beer, T.~Han, and P.~Hu, ``Draps: Dynamic and
  resource-aware placement scheme for docker containers in a heterogeneous
  cluster,'' in \emph{2017 IEEE 36th International Performance Computing and
  Communications Conference (IPCCC)}, Dec 2017, pp. 1--8.

\bibitem{joint:Hori}
T.~Hori \emph{et~al.}, ``Joint ctc/attention decoding for end-to-end speech
  recognition,'' in \emph{Annual Meeting of the Association for Computational
  Linguistics}, 2017.

\bibitem{lip:Wu}
D.~Wu and Q.~Ruan, ``Lip reading based on cascade feature extraction and hmm,''
  in \emph{International Conference on Signal Processing (ICSP)}, 2014.

\bibitem{lip:Morade}
S.~S. Morade and S.~Patnaik, ``Lip reading using dwt and lsda,'' in \emph{IEEE
  International Advance Computing Conference (IACC)}, 2014.

\bibitem{lip:Liu}
X.~Liu and Y.~m.~Cheung, ``Learning multi-boosted hmms for lip-password based
  speaker verification,'' in \emph{IEEE Transactions on Information Forensics
  and Security}, 2014.

\bibitem{lip:Zhao}
G.~Zhao \emph{et~al.}, ``Local spatiotemporal descriptors for visual
  recognition of spoken phrases,'' in \emph{Proceedings of the International
  Workshop on Human-centered Multimedia}, 2007.

\bibitem{svm:Shaikh}
A.~A. Shaikh \emph{et~al.}, ``Lip reading using optical flow and support vector
  machines,'' in \emph{International Congress on Image and Signal Processing},
  2010.

\bibitem{lip:Luettin}
J.~Luettin \emph{et~al.}, ``Visual speech recognition using active shape models
  and hidden markov models,'' in \emph{ICASSP}, 1996.

\bibitem{lip:Pale}
K.~Palecek, ``Comparison of depth-based features for lipreading,'' in
  \emph{International Conference on Telecommunications and Signal Processing
  (TSP)}, 2015.

\bibitem{AAM:Cootes}
T.~F. Cootes \emph{et~al.}, ``Active appearance models,'' \emph{IEEE
  Transactions on Pattern Analysis and Machine Intelligence}, vol.~23, no.~6,
  pp. 681--685, 2001.

\bibitem{lip:Potamianos}
G.~Potamianos \emph{et~al.}, ``An image transform approach for hmm based
  automatic lipreading,'' in \emph{ICIP}, 1998.

\bibitem{audiovisual:Sanabria}
R.~Sanabria \emph{et~al.}, ``{Robust end-to-end deep audiovisual speech
  recognition},'' \emph{arXiv:1611.06986}, 2016.

\bibitem{c3d:Tran}
D.~Tran \emph{et~al.}, ``Learning spatiotemporal features with 3d convolutional
  networks,'' in \emph{ICCV}, 2015.

\bibitem{highway:Srivastava}
R.~K. Srivastava \emph{et~al.}, ``Training very deep networks,'' in
  \emph{NIPS}, 2015.

\bibitem{lipread:Stafylakis}
T.~Stafylakis and G.~Tzimiropoulos, ``Combining residual networks with lstms
  for lipreading,'' in \emph{Interspeech}, 2017.

\bibitem{gru}
J.~Chung \emph{et~al.}, ``Empirical evaluation of gated recurrent neural
  networks on sequence modeling,'' \emph{arXiv:1412.3555}, 2014.

\bibitem{dlib:Davis}
D.~E. King, ``Dlib-ml: A machine learning toolkit,'' \emph{Journal of Machine
  Learning Research}, vol.~10, pp. 1755--1758, 2009.

\bibitem{papineni2002bleu}
K.~Papineni \emph{et~al.}, ``Bleu: a method for automatic evaluation of machine
  translation,'' in \emph{Proceedings of the 40th annual meeting on association
  for computational linguistics}, 2002.

\bibitem{tensorflow:Martin}
M.~Abadi \emph{et~al.}, ``Tensorflow: A system for large-scale machine
  learning,'' in \emph{USENIX Symposium on Operating Systems Design and
  Implementation (OSDI)}, 2016.

\bibitem{adam:Kingma}
D.~P. Kingma and J.~Ba, ``Adam: A method for stochastic optimization.'' in
  \emph{ICLR}, 2014.

\bibitem{lipread:Chung}
J.~S. Chung and A.~Zisserman, ``Lip reading in the wild,'' in \emph{ACCV},
  2016.

\bibitem{resnet:He}
K.~He \emph{et~al.}, ``Deep residual learning for image recognition,'' in
  \emph{CVPR}, 2016.

\bibitem{mapping:Neti}
P.~Neti \emph{et~al.}, ``Audio-visual speech recognition,'' in \emph{Technical
  report. Center for Language and Speech Processing, Johns Hopkins University,
  Baltimore}, 2000.

\end{thebibliography}

\end{document}